\documentclass[confidential,public]{innerlogic-technote}

\usepackage{xspace}
\usepackage{multirow}

\title{InnerLogic-Surg: An Open Surgical Foundation Model via Continual Pretraining}
\notenumber{IL-TN-[2026]-[11.1]}
\version{0.1}
\author{Kun Yuan, Hao Ding, Yiqing Shen, Gasper Podobnik, Benjamin Killeen, Tito Porras, Axel Krieger, Mathias Unberath}   
\affiliation{Inner Logic, Inc., Baltimore, MD}
\correspondence{kun@innerlogic.io}
\date{[November 1, 2026]}

\newcommand{\methodname}{InnerLogic-Surg\xspace}   

\begin{document}
\maketitle

\begin{abstract}
Vision foundation models are the backbone of modern surgical image analysis, yet most surgical systems still rely on encoders pretrained on natural images or on specialist models trained from scratch on comparatively small datasets. Pretraining encoders on large amounts of surgery-specific data has the potential to generally enhance the performance of surgical image analysis methods simply by providing stronger backbones. 
Here we present InnerLogic-Surg, an open family of surgical vision encoders (21M--300M parameters) obtained by continually pretraining a transformer-based image encoder on millions of surgical images curated from 11{,}965 internet-sourced surgical videos spanning $\sim$2{,}900 hours and more than fifteen procedure types. Our training recipe inherits DINOv3's pretraining stage to preserve representation ability after surgical adaptation, and additionally applies a learned feature-upsampling module for high-resolution dense prediction at low token cost. Across a benchmark covering workflow, instrument and anatomy recognition, segmentation, detection, and monocular depth, under the same architecture, the largest InnerLogic-Surg model achieves the best linear-probe performance in all 20 surgical tasks. 
We demonstrate that surgical continual pretraining yields transferable representations that generalize across procedures, anatomical regions, and task types, showing our surgical image encoders as a promising and scalable alternative to the procedure-specific representation learning approaches that have traditionally dominated surgical computer vision.

\end{abstract}

\section{Introduction}

Surgical computer vision has traditionally been built around task-specific models trained on carefully annotated datasets. This paradigm has produced strong systems for individual surgical tasks such as workflow recognition, instrument segmentation, and anatomy understanding. 
However, extending these models to new tasks, procedures, or institutions often requires additional annotations and retraining, making it difficult to scale this approach. 
At the same time, surgical video is generated at a scale far beyond what can realistically be labeled, leaving most of its visual information unused.

Foundation models seek to leverage this visual data at scale, and their success is driven largely by self-supervised learning (SSL), which enables reusable representations from vast collections of unlabeled images. Adapting such models to surgery, however, is not straightforward. Continued self-supervised training from general foundation models can erode spatial structure already present in the pretrained representations, creating a tension between acquiring surgical specificity and preserving the broad visual priors from general pretraining.

Recent work has begun to investigate surgery-specific foundation models. Self-supervised approaches, such as SurgeNetXL \cite{jaspers2025surgenet} and LemonFM \cite{che2025lemon}, leverage increasingly large datasets of endoscopic videos to train domain-specific visual encoders. Concurrently, multimodal surgical vision-language pretraining models, such as SurgVLP \cite{yuan2025surgvlp} and PeskaVLP \cite{yuan2024peskavlp}, exploit surgical lecture videos to align complex visual concepts with clinical text, showcasing powerful zero-shot capabilities. These models enable versatile downstream applications by fine-tuning massive architectures from scratch. However, they largely focus on learning representations within the surgical domain itself. Specifically, they omit the potential to continually adapt general-purpose foundation models to surgery while preserving the rich visual priors, generality, and spatial structure acquired during large-scale natural-image pretraining.

\begin{figure*}[t]
\centering
\begin{minipage}[t]{0.55\linewidth}
{\sffamily\bfseries a}\\[2pt]
\includegraphics[width=\linewidth]{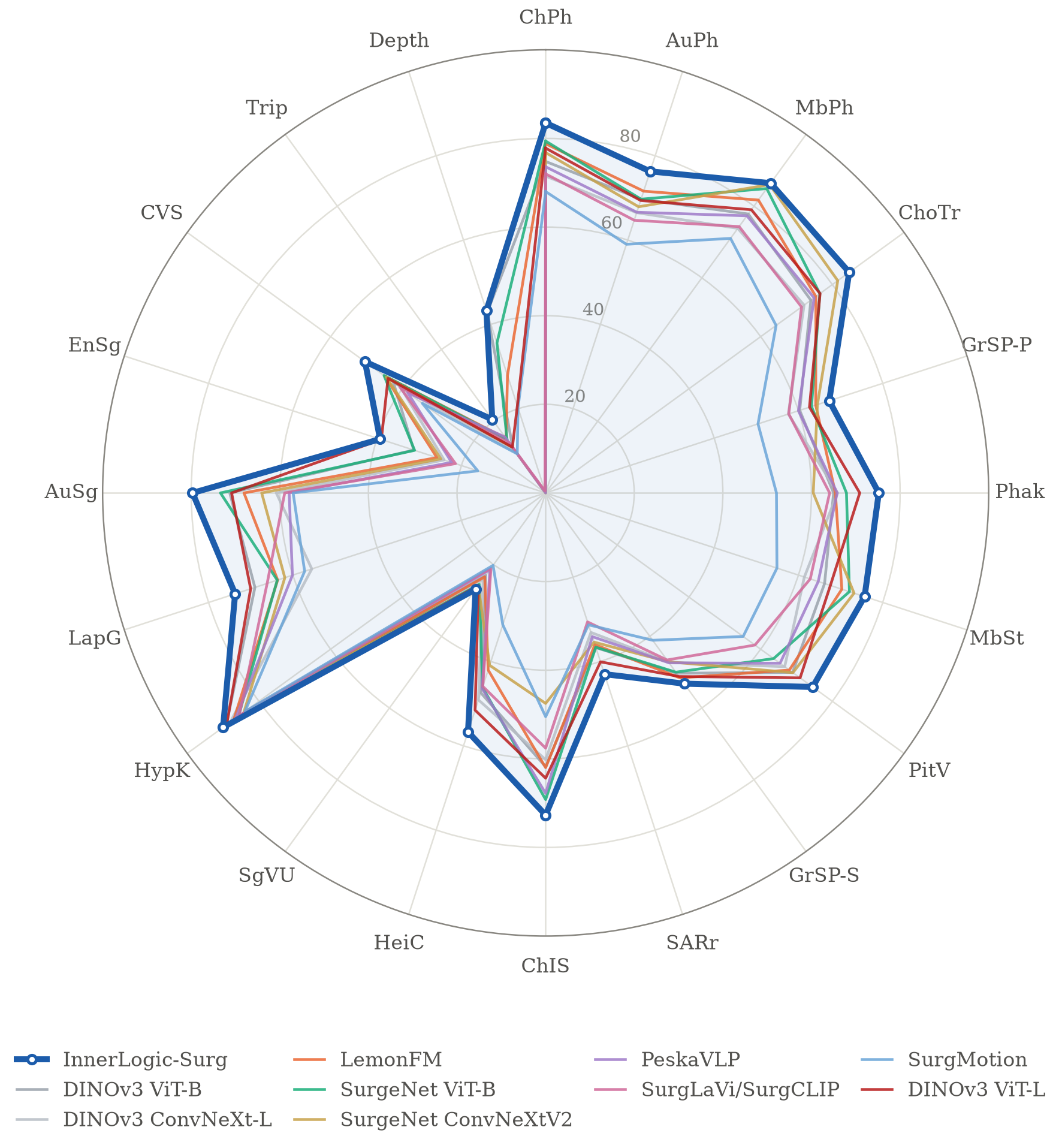}
\end{minipage}\hfill
\begin{minipage}[t]{0.41\linewidth}
{\sffamily\bfseries b}\\[2pt]
\includegraphics[width=\linewidth]{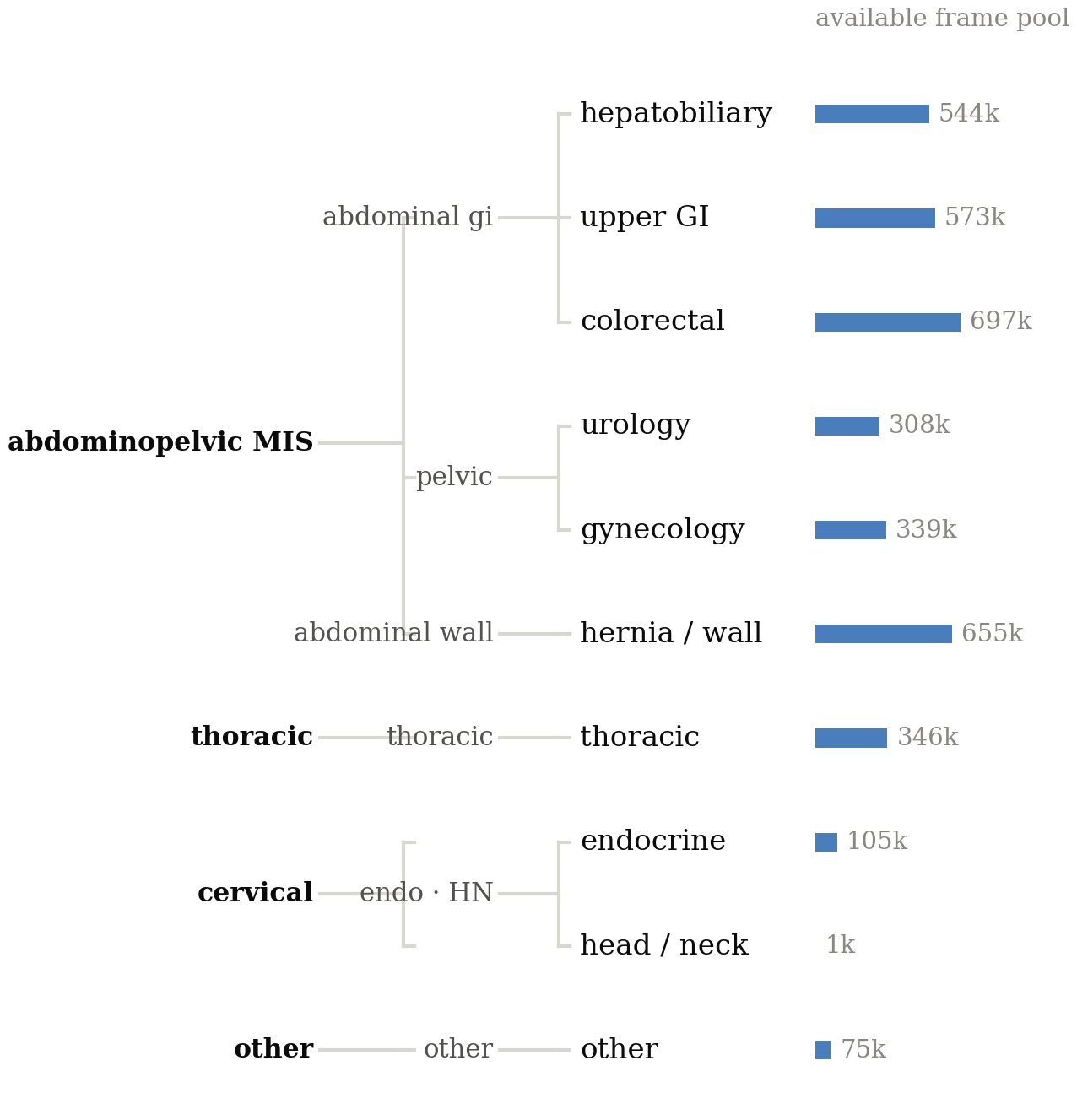}
\end{minipage}
\caption{\textbf{\methodname at a glance.}
\textbf{a}, Capability-grouped frozen linear-probe performance over 20 axes and 9 encoders, grouped by capability (phase, step, action,
instrument, and anatomy recognition; segmentation; detection; depth). Absolute scores (0--100); our
ViT-L (thick line) outperforms its counterparts on all 20 axes against all baseline surgical foundation models.
\textbf{b}, Specialty taxonomy of the pretraining corpus (operative field $\rightarrow$
parent $\rightarrow$ leaf) with each leaf's available frame pool: four operative fields and
more than fifteen procedure types, dominated by abdominopelvic minimally invasive surgery.}
\label{fig:overview}
\end{figure*}

Here we introduce InnerLogic-Surg, an open family of surgical foundation models designed as the next step towards realizing the vision of scalable, generalizable surgical AI. We curate a large and diverse corpus of publicly available surgical video spanning procedures, anatomies, and acquisition settings (\cref{fig:overview}b). We further build upon DINOv3\cite{oquab2025dinov3} to inherit broad, general-purpose visual knowledge while rapidly adapting to the surgical domain, designed to counteract the degradation of dense feature maps that otherwise afflicts long domain adaptations. To support dense surgical prediction, InnerLogic-Surg further adopts a specialized feature super-resolution module that produces high-resolution feature maps without increasing the backbone token resolution via AnyUP \cite{wimmer2026anyup}.

The result is a family of surgical vision encoders spanning 21M to 300M parameters, designed to provide strong transferable representations across a range of computational budgets. Under a frozen-backbone evaluation protocol, our vision foundation model outperforms its counterparts of the same architecture, i.e., at matched architecture (ViT-L/16 against ViT-L/16 and ViT-B/16 against ViT-B/16, with identical depth, width and patch size), across a broad range of benchmarks spanning six capability groups (\cref{fig:overview}a), with the largest gains on dense prediction tasks, including segmentation and depth estimation, and so on.

\section{Results}

\subsection{A unified model family leads across surgical benchmark axes}
\label{sec:downstream}

\begin{table*}[t]
\centering
\footnotesize
\setlength{\tabcolsep}{2pt}

\caption{\textbf{Downstream linear-probe evaluation across the full 20-task benchmark} (\cref{tab:benchmark}), which defines every abbreviation, metric, and test-set size. All
recognition, segmentation, CVS, and Trip scores are reported $\times$100;
higher is better for these metrics (accuracy for recognition columns ChPh--LapG and
ChoTr--Phak, mIoU for AuSg and EnSg, mAP for CVS, and top-20 tmAP for Trip).
Depth is reported as SI-RMSE (lower is better, not multiplied by 100); Params are
nominal backbone parameter counts; Video action
recognition is reported separately in the text. Our flagship is the ViT-L configuration; the compact ViT-B variant
is included for the efficiency comparison. AnyUp is not used in any number here.}
\label{tab:linear_probe}

\resizebox{\textwidth}{!}{%
\begin{tabular}{llr*{20}{c}}
\toprule
Model & Arch. & Params & ChPh & AuPh & MbPh & MbSt & PitV & HypK & ChIS & HeiC & SARr & SgVU & LapG & AuSg & EnSg & CVS & Trip & ChoTr & GrSP-P & GrSP-S & Phak & Depth\,$\downarrow$ \\
\midrule

\multirow{3}{*}{DINOv3 (LVD)}
& ViT-B & 86M & 74.8 & 69.7 & 77.8 & 66.2 & 68.9 & 89.1 & 61.8 & 47.4 & 36.7 & 23.0 & 69.0 & 71.3 & 35.1 & 42.5 & 12.6 & 74.0 & 60.3 & 50.3 & 65.0 & 0.318 \\
& ViT-L & 300M & 77.8 & 69.4 & 79.0 & 67.6 & 71.0 & 89.0 & 64.4 & 51.5 & 40.1 & 25.8 & 70.0 & 70.9 & 39.0 & 44.0 & 12.8 & 76.6 & 62.6 & 51.2 & 70.9 & 0.364 \\
& ConvNeXt-L & 198M & 71.5 & 66.5 & 73.8 & 61.4 & 66.7 & 88.0 & 60.2 & 49.1 & 33.1 & 24.0 & 55.5 & 60.8 & 25.9 & 40.4 & 11.3 & 72.1 & 57.6 & 47.2 & 65.9 & 0.526 \\

LemonFM & ConvNeXt-L & 198M & 78.8 & 71.6 & 81.7 & 70.3 & 68.0 & 87.0 & 62.0 & 42.0 & 36.0 & 23.4 & 63.6 & 68.1 & 28.0 & 44.9 & 15.2 & 75.4 & 64.1 & 51.6 & 65.4 & 0.349 \\

\multirow{2}{*}{SurgeNet}
& ViT-B & 86M & 79.4 & 69.7 & 84.9 & 72.1 & 63.6 & 85.7 & 69.2 & 46.1 & 36.6 & 25.5 & 63.7 & 73.4 & 32.3 & 45.1 & 14.8 & 76.5 & 63.1 & 50.0 & 67.9 & 0.333 \\
& ConvNeXtV2-T & 29M & 76.8 & 67.9 & 86.0 & 73.2 & 69.0 & 84.2 & 47.5 & 40.9 & 35.4 & 25.9 & 61.9 & 64.1 & 26.5 & 44.5 & 12.7 & 81.5 & 64.5 & 47.1 & 60.4 & 0.438 \\

PeskaVLP & ResNet-50 & 26M & 73.6 & 66.6 & 77.3 & 64.7 & 65.4 & 86.5 & 67.8 & 46.4 & 34.1 & 21.4 & 60.1 & 57.9 & 23.8 & 38.5 & 15.6 & 74.9 & 60.0 & 47.4 & 65.8 & 0.492 \\

SurgMotion & ViT-L & 300M & 68.0 & 59.0 & 71.0 & 54.9 & 55.1 & 83.9 & 50.5 & 31.3 & 31.3 & 20.2 & 57.2 & 57.0 & 20.6 & 34.4 & 10.9 & 64.3 & 50.4 & 41.1 & 52.1 & 0.365 \\

SurgLaVi/SurgCLIP & TimeSformer-B & 121M & 72.0 & 64.7 & 74.3 & 62.8 & 58.4 & 85.5 & 57.6 & 45.9 & 30.6 & 21.0 & 66.1 & 58.9 & 22.9 & 40.7 & 14.0 & 71.4 & 57.7 & 46.6 & 64.1 & 0.406 \\

\midrule

\multirow{2}{*}{InnerLogic-Surg (ours)}
& ViT-B & 86M & 80.7 & 72.9 & 84.9 & 74.0 & \textbf{74.5} & 88.9 & 67.0 & 50.7 & 40.8 & 23.9 & 72.1 & 76.0 & 37.1 & 43.7 & 17.1 & 78.4 & 65.9 & \textbf{53.5} & 70.1 & \textbf{0.293} \\
& ViT-L & 300M & \textbf{83.5} & \textbf{76.3} & \textbf{86.4} & \textbf{75.7} & \textbf{74.5} & \textbf{90.0} & \textbf{72.7} & \textbf{56.7} & \textbf{43.0} & \textbf{26.9} & \textbf{73.7} & \textbf{79.7} & \textbf{39.3} & \textbf{50.4} & \textbf{20.5} & \textbf{84.7} & \textbf{67.3} & 53.2 & \textbf{75.2} & 0.317 \\

\bottomrule
\end{tabular}%
}
\end{table*}

\begin{table*}[t]
\centering
\footnotesize
\setlength{\tabcolsep}{4pt}
\caption{\textbf{The 20-task benchmark.} Every dataset, its abbreviation as used in
\cref{tab:linear_probe} and \cref{fig:overview}a, task, evaluation metric, imaging modality,
procedure domain, class count, and held-out test-set size (frames). All probes are linear
heads on frozen features (a $1{\times}1$-conv head for depth); splits are by video. All
metrics are higher-is-better except SI-RMSE ($\downarrow$); the depth test set is the
20 held-out SCARED keyframes.}
\label{tab:benchmark}

\resizebox{\textwidth}{!}{%
\begin{tabular}{llllllrr}
\toprule
Abbr. & Dataset & Task & Metric & Modality & Procedure domain & Classes & Test size \\
\midrule
ChPh & Cholec80\cite{twinanda2016endonet} & Phase recognition & Acc & Laparoscopy & Cholecystectomy & 7 & 98{,}234 \\
AuPh & AutoLaparo\cite{wang2022autolaparo} & Phase recognition & Acc & Laparoscopy & Hysterectomy & 7 & 28{,}060 \\
MbPh & MultiBypass140\cite{lavanchy2024challenges} & Phase recognition & Acc & Laparoscopy & Gastric bypass & 12 & 231{,}303 \\
MbSt & MultiBypass140\cite{lavanchy2024challenges} & Step recognition & Acc & Laparoscopy & Gastric bypass & 46 & 138{,}324 \\
PitV & PitVis\cite{das2024pitvis} & Step recognition & Acc & Endonasal endoscopy & Pituitary adenoma resection & 15 & 19{,}017 \\
HypK & HyperKvasir\cite{borgli2020hyperkvasir} & Findings classification & Acc & Flexible GI endoscopy & Gastro-/colonoscopy & 23 & 1{,}610 \\
ChIS & CholecInstanceSeg\cite{alabi2025cholecinstanceseg} & Instrument presence & Acc & Laparoscopy & Cholecystectomy & 7 & 1{,}241 \\
HeiC & HeiChole\cite{wagner2023comparative} & Instrument presence & Acc & Laparoscopy & Cholecystectomy & 7 & 1{,}845 \\
SARr & SAR-RARP50\cite{psychogyios2023sar} & Gesture recognition & Acc & Robotic & Prostatectomy (suturing) & 7 & 1{,}986 \\
SgVU & SurgVU\cite{zia2023surgical} & Instrument presence & Acc & Robotic & Multi-procedure (da Vinci) & 20 & 2{,}778 \\
LapG & LapGC-KVAD-30 & Anatomy recognition & Acc & Laparoscopy & Gastric cancer surgery & 15 & 1{,}317 \\
AuSg & AutoLaparo (Task 3)\cite{wang2022autolaparo} & Semantic segmentation & mIoU & Laparoscopy & Hysterectomy & 4 & 438 \\
EnSg & EndoVis18\cite{endovis18} & Scene segmentation & mIoU & Robotic & Porcine nephrectomy & 12 & 449 \\
CVS & Endoscapes-CVS\cite{murali2023endoscapes} & Critical view of safety & mAP & Laparoscopy & Cholecystectomy & 3 & 1{,}799 \\
Trip & CholecT50\cite{nwoye2022rendezvous} & Action triplets & tmAP & Laparoscopy & Cholecystectomy & 100 & 5{,}500 \\
ChoTr & CholecTrack20\cite{nwoye2023cholectrack20} & Phase recognition & Acc & Laparoscopy & Cholecystectomy & 7 & 6{,}226 \\
GrSP-P & GraSP\cite{ayobi2025pixel} & Phase recognition & Acc & Robotic & Prostatectomy & 11 & 53{,}444 \\
GrSP-S & GraSP\cite{ayobi2025pixel} & Step recognition & Acc & Robotic & Prostatectomy & 21 & 53{,}444 \\
Phak & PhaKIR\cite{rueckert2024phakir} & Phase recognition & Acc & Laparoscopy & Cholecystectomy & 8 & 5{,}382 \\
\midrule
Depth & SCARED\cite{allan2021stereo} & Monocular depth (3D) & SI-RMSE\,$\downarrow$ & Stereo laparoscopy & Porcine abdomen (da~Vinci) & --- & 20 \\
\bottomrule
\end{tabular}%
}
\end{table*}

We evaluate the \methodname family by linear probing on frozen
backbone features. We train a linear layer for each task with no deep parameterized
fine-tuning. The performance reflects the raw quality of the self-supervised
representations (Methods). We compare against the general-purpose DINOv3
family\cite{oquab2025dinov3} and the specialized surgical encoders
PeskaVLP\cite{yuan2024peskavlp}, LemonFM\cite{che2025lemon},
SurgeNet\cite{de2026towards}, SurgMotion\cite{wu2026surgmotion}, and
SurgLaVi\cite{perez2026surglavi}, on a benchmark of 20 tasks in six capability
groups: workflow recognition (phase, step, gesture) on
Cholec80\cite{twinanda2016endonet}, AutoLaparo\cite{wang2022autolaparo},
MultiBypass140\cite{lavanchy2024challenges},
CholecTrack20\cite{nwoye2023cholectrack20}, GraSP\cite{ayobi2025pixel},
SAR-RARP50\cite{psychogyios2023sar} and others; instrument recognition,
including CholecInstanceSeg\cite{alabi2025cholecinstanceseg} and
HeiChole\cite{wagner2023comparative}; anatomy and findings recognition; dense
semantic segmentation; structured multi-label prediction on
Endoscapes-CVS\cite{murali2023endoscapes} and CholecT50
triplets\cite{nwoye2022rendezvous}; and monocular depth on
SCARED\cite{allan2021stereo}. We evaluate temporal video understanding
separately on SurgicalActions160.

As shown in \cref{tab:linear_probe}, our continually pretrained \methodname
family obtains the strongest result on each of the 20 benchmark axes among
the nine evaluated baseline configurations (19 image-based tasks plus depth;
video recognition is reported separately below). \Cref{fig:overview}a
summarizes this capability profile. The flagship ViT-L leads on 18 axes,
while the compact ViT-B achieves the best GraSP step-recognition and
monocular-depth results; both variants tie on PitVis step recognition.
Despite having only 86M parameters, our ViT-B outperforms the 198M-parameter
LemonFM ConvNeXt-Large on 19 of the 20 benchmark axes
(\cref{tab:linear_probe}).

\paragraph{Workflow recognition (phase, step, action)}
Workflow recognition characterizes the progression of a surgical procedure by assigning semantic workflow labels, such as phase, step, or action, to individual video frames or time points. Across the six phase axes (ChPh, AuPh, MbPh, ChoTr, GrSP-P, Phak), the three
step axes (MbSt, PitV, GrSP-S), and action recognition (SARr), \methodname
improves on the strongest baseline of each task by $+3.0$ on average, e.g.\ Cholec80
phase $83.5$ against $79.4$ for the best surgical baseline, and CholecTrack20
$84.7$ against $81.5$. These are the axes prior surgical encoders were built
for; ours leads on every one, with per-task margins from $+0.4$ (MbPh) to
$+4.7$ (AuPh).

\paragraph{Instrument and anatomy recognition}
Instrument and anatomy recognition evaluates whether a visual representation can identify the surgical tools and anatomical structures present in individual frames, often under substantial appearance variation, occlusion, and visual similarity between classes. Across the instrument-centric axes (ChIS, HeiC, SgVU) and anatomy axes (HypK, LapG), \methodname consistently outperforms the strongest baseline, with particularly large gains on more fine-grained recognition tasks. For example, it reaches $56.7$ on HeiC compared with $51.5$, $72.7$ on ChIS compared with $69.2$, and $73.7$ on LapG compared with $70.0$.

\paragraph{Structured multi-label prediction}
These tasks test whether the representation captures more than coarse scene
semantics: Endoscapes-CVS assesses the anatomical criteria for the critical
view of safety, while CholecT50 evaluates instrument--action--target triplets.
\methodname reaches $50.4$ mAP for CVS compared with $45.1$ for the
strongest baseline ($+5.3$ points), and $20.5$ top-20 tmAP for triplets
compared with $15.6$ ($+4.9$ points).


\paragraph{Video action recognition (SurgicalActions160)}
Our frame-level foundation encoder also supports temporal recognition at the clip level. SurgicalActions160 \cite{schoeffmann2018video} contains 160 clips across 16 action classes. Each frozen encoder yields a per-frame global feature, averaged over the frames of each clip and L2-normalized. A linear head is trained with Adam for 200 full-batch steps on 128 clips and tested on 32 (two per class), with the learning rate selected on training accuracy. InnerLogic-Surg ViT-L reaches 53.1\% top-1 and 49.6 macro-F1 without any video architecture or video pretraining. It is comparable to LemonFM (53.1, 50.2), a ConvNeXt-L, and to DINOv3 ViT-L (56.3, 51.7), and trails SurgLaVi/SurgCLIP (59.4, 56.7), a TimeSformer-B video model, by 6.3 points.

\begin{figure*}[tp]
\centering
\includegraphics[width=\textwidth]{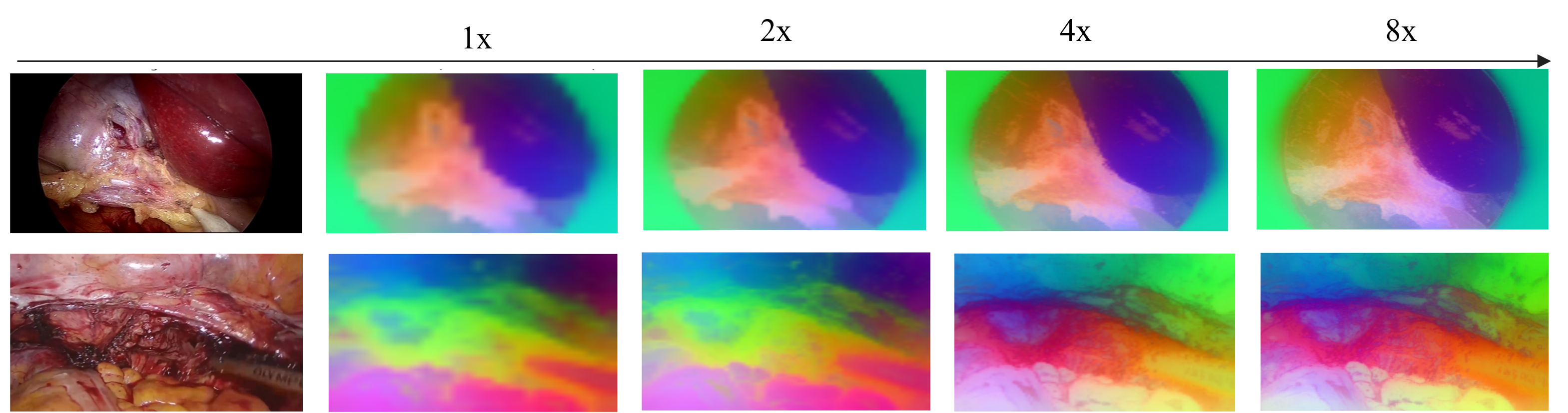}
\caption{\textbf{Dense feature maps.} PCA of \methodname patch
features for two surgical frames. We show the feature map at
$\times1$, $\times2$, $\times4$ and $\times8$ resolution, each produced by lifting the
patch features with the applied upsampling module\cite{wimmer2026anyup}. At $\times1$ the map is a coarse patch grid with blocky
edges. As the resolution grows, the same regions are traced at a finer scale, and by
$\times4$ and $\times8$ the boundaries of the liver, fat and instruments are sharp
and fine tissue texture is visible.}
\label{fig:pca}
\end{figure*}

\begin{figure*}[tp]
\centering
{\sffamily\bfseries a}\\[2pt]
\includegraphics[width=\textwidth]{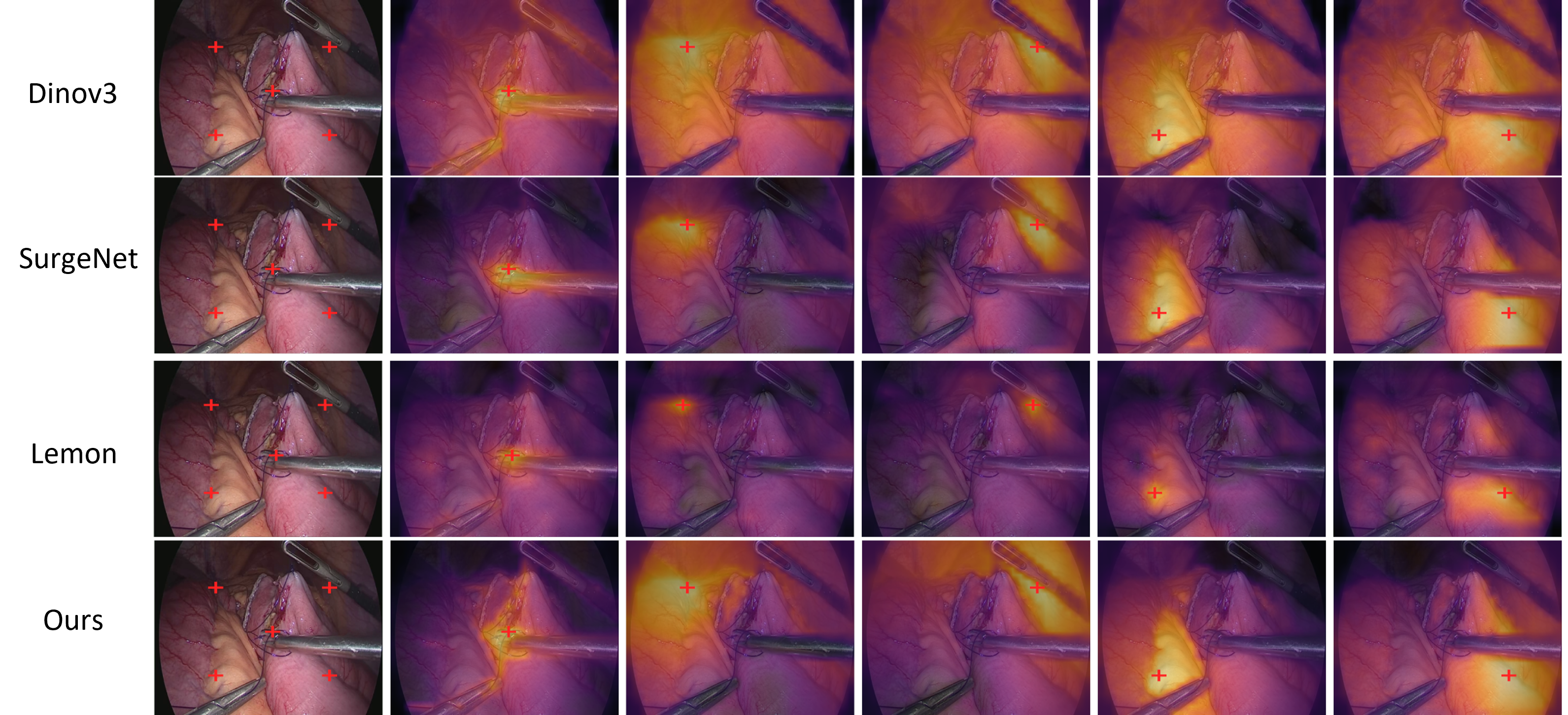}\\[8pt]
{\sffamily\bfseries b}\\[2pt]
\includegraphics[width=\textwidth]{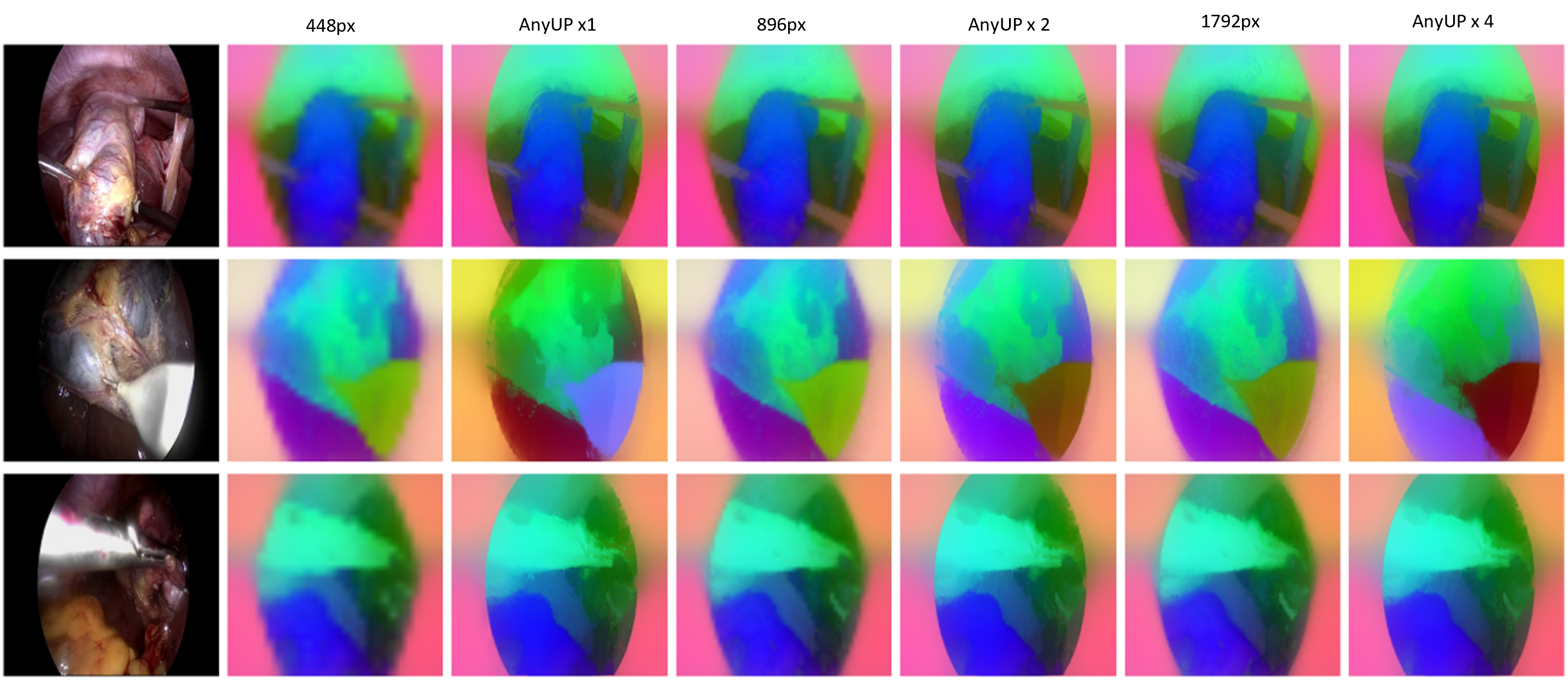}
\caption{\textbf{What protects dense features.}
\textbf{a}, our foundation model preserves sharp, highly localized
spatial features across the surgical field.
\textbf{b}, High-resolution feature maps at $\times1$, $\times2$, $\times4$ input
resolution. At each scale, patch features from high-resolution input (bilinearly upsampled)
against the same features lifted to pixel resolution by the learned image-guided upsampler
(AnyUp), which retains sharper anatomical and tool boundaries.}
\label{fig:denseq}
\end{figure*}

\subsection{Gains on dense prediction and 3D understanding}
\label{sec:dense3d}

Previous evaluations of surgical foundation models have primarily focused on coarse-grained semantic understanding, often relying on frame-level classification benchmarks.
However, practical surgical applications require more than recognizing what is happening in an image. They also depend on fine-grained spatial understanding and 3D perception, including segmentation, depth estimation, and tracking.
To assess these capabilities, we expand the evaluation to a broader range of surgical perception tasks and show that continual pretraining provides substantial gains.

\paragraph{Dense prediction (segmentation)}
Segmentation evaluates whether the representation preserves fine-grained
spatial information by assigning a semantic label to each pixel or local
region. On AutoLaparo (AuSg), \methodname ViT-L reaches $79.7$ mIoU,
compared with $73.4$ for the strongest baseline ($+6.3$ points). On
EndoVis18 (EnSg), it achieves $39.3$ mIoU versus $39.0$ for DINOv3
ViT-L ($+0.3$ points). Thus, the gains extend to both segmentation
datasets, although the improvement is substantially larger on AutoLaparo.
Unlike frame-level classification, these dense probes directly assess
patch-level feature quality needed to distinguish fine spatial boundaries.

\paragraph{3D understanding (monocular depth)}
Monocular depth estimation provides a direct measure of how well visual representations capture scene geometry from a single RGB image. On SCARED, at matched architecture, \methodname achieves a lower SI-RMSE than geometry-oriented DINOv3-LVD for ViT-L ($0.317, 0.364$) and ViT-B variants ($0.293,  0.318$). Notably, our ViT-B variant achieves the best performance among all evaluated encoders, 86M-parameter. In contrast, the ConvNeXt-L, PeskaVLP and SurgeNet-ConvNeXtV2 encoders do not improve on a simple constant-depth baseline ($0.407$), indicating that their learned features contain limited information about scene geometry; LemonFM ($0.349$) is the only convolutional encoder that does. We use SI-RMSE rather than the conventional $\delta<1.25$ metric because SCARED's narrow within-frame depth range allows a constant-depth predictor to perform artificially well under the latter (Methods). These results suggest that semantic recognition performance alone does not ensure geometric accuracy; coarse spatial feature maps in some convolutional encoders may further limit depth prediction.

\paragraph{Dense features, visualized (\cref{fig:pca})} 
We further project the frozen patch features of each encoder onto their top principal
components and render them as color maps. \methodname learns to distinguish meaningful structures within the surgical scene without explicit supervision. Instruments, individual organs, and surrounding anatomy form distinct, spatially coherent clusters, with boundaries that closely align with the actual instrument-tissue interfaces. In contrast, previous state-of-the-art methods produce coarser scene representations with less clearly defined boundaries. LemonFM, for example, is a ConvNeXt that yields a $7\times7$ map at our 224-pixel input, whereas our ViT-L/16 yields $14\times14$. The coarser grid may limit fine spatial detail and contribute to LemonFM's gaps in segmentation and monocular depth estimation.

\paragraph{Dense feature visualization (\cref{fig:denseq})}
\cref{fig:denseq}a shows our model's feature responses around selected reference points on surgical frames. The responses stay localized across the surgical field.

There are multiple ways of achieving pixel-level resolution with high
quality: run the backbone at higher input resolution, or lift the coarse patch grid (14$\times$14 at our 224-pixel protocol, 28$\times$28 at 448-pixel input) to pixel resolution with a learned, image-guided
feature upsampler. We use AnyUp~\cite{wimmer2026anyup} for the latter, applying it
directly to our frozen backbone's patch features.
\Cref{fig:denseq}b contrasts the two approaches at $\times1$/$\times2$/$\times4$
scales on the same frames: naive high-resolution inference introduces the
grid artifacts and spatial drift typical of long-context ViT-S, while
AnyUp-upsampled features retain crisp anatomical and tool contours at every
scale, without any extra backbone passes.

\section{Methods}

\subsection{Continual training with self-supervision}

To effectively adapt general-purpose visual representations to the surgical domain, we initialize our model using the DINOv3 framework \cite{oquab2025dinov3} and perform continual pretraining on our curated 3.6 million frames. This pretraining corpus is decontaminated against the evaluation test sets. Our training pipeline optimizes a joint self-supervised objective, combining the DINO global image-level loss with the iBOT masked patch-level loss. This ensures the model simultaneously learns invariant, global surgical scene context and fine-grained, localized tool-tissue interactions. A second training stage adds the Gram-anchoring objective, which constrains the pairwise patch-feature similarities (the Gram matrix) to remain close to those of an earlier checkpoint (loss curves for all objectives in Extended Data Fig.~\ref{edfig:loss}). We adopt this stage as part of the recipe; its downstream contribution is not individually ablated in this report.

\paragraph{Lightweight backbone for surgical deployment}
Our \methodname models are built upon highly efficient Vision Transformer backbones; we describe the ViT-B/16 reference configuration here, with only 85.7 million parameters, a depth of 12, and an embedding dimension of 768. While this lightweight architecture sacrifices the theoretical upper limit of representation capacity compared to much larger 7B parameter frontier models like DINOv3-7B, we deliberately accept this cost. The exorbitant latency of giant models renders them impractical for dense surgical video tasks. By leveraging our surgical pretraining strategy, we extract high-fidelity spatial representations from the ViT-B architecture without incurring the clinical deployment bottlenecks of heavier architectures.

\paragraph{High-resolution extrapolation} Naively running the backbone at
resolutions beyond its pretraining crop size degrades the patch-level feature
map, generating grid artifacts since token geometry at test time is governed by a pathway
the model never saw outside a fixed crop range during training. In this work, we achieve any resolution with AnyUp~\cite{wimmer2026anyup}, a universal
feature upsampler trained once, independently of the backbone, that turns the
backbone's native-resolution patch tokens into dense features that stay
localized and sharply delineated across the surgical field rather than
spreading diffusely across tissue. Concurrent work on surgical foundation
models addresses the same degradation differently: SURGE\cite{stilz2026surge} folds an additional
post-training refinement phase directly into the backbone, i.e., 30k further
iterations of multi-crop training across several resolutions together with a
Gram-anchoring loss that aligns student/teacher patch-token Gram
matrices, which bakes resolution robustness into the encoder weights at the
cost of extra continual-pretraining compute tied to that one checkpoint.
Our foundation model efficiently pairs RoPE with a backbone-agnostic upsampler to address the two
concerns: the pretraining recipe is unchanged, and the dense-feature fix can
be swapped or improved independently of any future re-training of the
encoder.

\paragraph{Vision transformers and high-capacity projection heads}
We adopt Vision Transformers (ViTs) rather than conventional convolutional backbones such as ResNet and ConvNeXt to support both global semantic understanding and fine-grained spatial perception. The patch-based architecture preserves spatially organized feature representations, while self-attention captures long-range relationships between instruments, tissues, and anatomical structures. This design is particularly suited to dense prediction tasks such as segmentation and depth estimation. To investigate the effect of model capacity, we develop a family of surgical image encoders spanning ViT-S, ViT-B, and ViT-L, providing a consistent architecture across different computational budgets. In addition, we adopt an asymmetric training architecture with high-capacity projection heads. Rather than scaling the projection heads proportionally to the backbone, we retain the large heads used in DINOv3-7B across our model family. This provides substantial capacity for self-supervised learning without increasing the inference cost of the backbone. Extended Data Fig.~\ref{edfig:loss} shows rapid and stable convergence throughout continual pretraining, further supporting this design choice.

\begin{itemize}[leftmargin=*]
    \item \textbf{DINO Head:} We employ an 8192-dimensional hidden layer with a 512-dimensional bottleneck and an extensive array of 65,536 prototypes.
    \item \textbf{iBOT Head:} We utilize a 4096-dimensional hidden layer with a 384-dimensional bottleneck and 65,536 prototypes.
\end{itemize}

\subsection{Downstream evaluation protocol}
We score our models with two separate linear-probe setups. The main one, used for our headline
results, trains a simple linear readout on top of frozen features across our
full set of downstream tasks. 

Classification probes share one protocol. Training images use RandomResizedCrop to 224 with horizontal flips; evaluation images are resized to 256 and center-cropped to 224. ViT features are the CLS token and the average-pooled final-block patch tokens; ConvNeXt and ResNet features are the global average pool of the final stage. Each probe is a linear layer trained with SGD (momentum 0.9, no weight decay, cosine schedule) for 10 epochs of 1,250 iterations at batch size 256, over a 13-point learning-rate grid from 1e-5 to 1e-1. This applies equally to every encoder in the table.

For the ablation study, we switch to a lighter version, with the features computed once
and cached, and a linear readout is trained briefly
on top. The two setups don't agree in absolute
terms, and the gap isn't the same from task to task, which is exactly why
we never place their results side by side. The lighter setup also isn't
perfectly repeatable: rerunning it on the same checkpoint gives slightly
different numbers each time, usually by a small margin and occasionally a
bit more.

Two tasks sit outside the above protocols. For depth, we train a linear layer to
predict depth on the SCARED benchmark from frozen features. We skip the
usual accuracy-within-threshold metric here: the depth range in these frames
is so narrow that a model which just guesses one constant depth for
everything would win on that metric, making it useless for telling models
apart. We instead score depth with a standard scale-invariant error measure. For
video, a linear layer on frozen frame features classifies short surgical
action clips, evaluated against a held-out clip set where random guessing
would rarely succeed.

\subsection{Statistics}
The fourteen ablation tasks vary substantially in test-set size, which affects how reliably differences between pretraining configurations can be interpreted. To account for this, we compare the accuracy range across five equal-volume data configurations with twice the frame-level binomial standard error for each task (Extended Data Fig.~\ref{edfig:resolution}). Eight tasks exceed this threshold, while the remaining six show differences too small to reliably distinguish between configurations under this criterion. Because frames from the same video are correlated, these estimates may underestimate the true uncertainty and should be interpreted cautiously. We separately report the improvements from surgical continual pretraining over the original DINOv3 initialization in Extended Data Fig.~\ref{edfig:gain}.

\subsection{Code availability}
Training, curation, and evaluation code, together with model weights for the
released ViT-S/16, ViT-B/16, and ViT-L/16 encoders, will be made available at
the project page: \url{https://github.com/semaphor-surgical/Innerlogic-Surg}

\subsection{Statistics}
The fourteen ablation tasks vary substantially in test-set size, which affects how reliably differences between pretraining configurations can be interpreted. To account for this, we compare the accuracy range across five equal-volume data configurations with twice the frame-level binomial standard error for each task (Extended Data Fig.~\ref{edfig:resolution}). Eight tasks exceed this threshold, while the remaining six show differences too small to reliably distinguish between configurations under this criterion. Because frames from the same video are correlated, these estimates may underestimate the true uncertainty and should be interpreted cautiously. We separately report the improvements from surgical continual pretraining over the original DINOv3 initialization in Extended Data Fig.~\ref{edfig:gain}.

\subsection{Code availability}
Training, curation, and evaluation code, together with model weights for the
released ViT-S/16, ViT-B/16, and ViT-L/16 encoders, will be made available at
the project page.

\bibliographystyle{unsrtnat}
\bibliography{references}

\begin{thebibliography}{27}
\providecommand{\natexlab}[1]{#1}
\providecommand{\url}[1]{\texttt{#1}}
\expandafter\ifx\csname urlstyle\endcsname\relax
  \providecommand{\doi}[1]{doi: #1}\else
  \providecommand{\doi}{doi: \begingroup \urlstyle{rm}\Url}\fi

\bibitem[Jaspers et~al.(2025)]{jaspers2025surgenet}
Tim Jaspers et~al.
\newblock Scaling up self-supervised learning for improved surgical foundation models.
\newblock \emph{arXiv preprint arXiv:2501.09436}, 2025.

\bibitem[Che et~al.(2026)]{che2025lemon}
Chengan Che et~al.
\newblock {LEMON}: A large endoscopic monocular dataset and foundation model for perception in surgical settings.
\newblock In \emph{Proceedings of the IEEE/CVF Conference on Computer Vision and Pattern Recognition (CVPR)}, 2026.

\bibitem[Yuan et~al.(2025)]{yuan2025surgvlp}
Kun Yuan et~al.
\newblock Learning multi-modal representations by watching hundreds of surgical video lectures.
\newblock \emph{Medical Image Analysis}, 2025.

\bibitem[Yuan et~al.(2024)]{yuan2024peskavlp}
Kun Yuan et~al.
\newblock Procedure-aware surgical video-language pretraining with hierarchical knowledge augmentation.
\newblock In \emph{Advances in Neural Information Processing Systems (NeurIPS)}, 2024.

\bibitem[Oquab et~al.(2025)]{oquab2025dinov3}
Maxime Oquab et~al.
\newblock {DINOv3}: Self-supervised learning at scale.
\newblock Technical report, Technical Report, 2025.

\bibitem[Wimmer et~al.(2026)Wimmer, Truong, Rakotosaona, Oechsle, Tombari, Schiele, and Lenssen]{wimmer2026anyup}
Thomas Wimmer, Prune Truong, Marie-Julie Rakotosaona, Michael Oechsle, Federico Tombari, Bernt Schiele, and Jan~Eric Lenssen.
\newblock Anyup: Universal feature upsampling.
\newblock In \emph{Proceedings of the International Conference on Learning Representations ({ICLR})}, 2026.

\bibitem[Twinanda et~al.(2017)Twinanda, Shehata, Mutter, Marescaux, De~Mathelin, and Padoy]{twinanda2016endonet}
Andru~P Twinanda, Sherif Shehata, Didier Mutter, Jacques Marescaux, Michel De~Mathelin, and Nicolas Padoy.
\newblock Endonet: a deep architecture for recognition tasks on laparoscopic videos.
\newblock \emph{IEEE Transactions on Medical Imaging}, 36\penalty0 (1):\penalty0 86--97, 2017.

\bibitem[Wang et~al.(2022)Wang, Lu, Long, Zhong, Cheung, Dou, and Liu]{wang2022autolaparo}
Ziyi Wang, Bo~Lu, Yonghao Long, Fangxun Zhong, Tak-Hong Cheung, Qi~Dou, and Yunhui Liu.
\newblock Autolaparo: A new dataset of integrated multi-tasks for image-guided surgical automation in laparoscopic hysterectomy.
\newblock In \emph{International Conference on Medical Image Computing and Computer-Assisted Intervention}, pages 486--496. Springer, 2022.

\bibitem[Lavanchy et~al.(2024)Lavanchy, Ramesh, Dall’Alba, Gonzalez, Fiorini, M{\"u}ller-Stich, Nett, Marescaux, Mutter, and Padoy]{lavanchy2024challenges}
Jo{\"e}l~L Lavanchy, Sanat Ramesh, Diego Dall’Alba, Cristians Gonzalez, Paolo Fiorini, Beat~P M{\"u}ller-Stich, Philipp~C Nett, Jacques Marescaux, Didier Mutter, and Nicolas Padoy.
\newblock Challenges in multi-centric generalization: phase and step recognition in roux-en-y gastric bypass surgery.
\newblock \emph{International journal of computer assisted radiology and surgery}, 19\penalty0 (11):\penalty0 2249--2257, 2024.

\bibitem[Das et~al.(2024)Das, Khan, Hanrahan, Marcus, Stoyanov, et~al.]{das2024pitvis}
Adrito Das, Danyal~Z Khan, John~G Hanrahan, Hani~J Marcus, Danail Stoyanov, et~al.
\newblock {PitVis-2023} challenge: Workflow recognition in videos of endoscopic pituitary surgery.
\newblock \emph{arXiv preprint arXiv:2409.01184}, 2024.

\bibitem[Borgli et~al.(2020)Borgli, Thambawita, Smedsrud, Hicks, Jha, et~al.]{borgli2020hyperkvasir}
Hanna Borgli, Vajira Thambawita, Pia~H Smedsrud, Steven Hicks, Debesh Jha, et~al.
\newblock {HyperKvasir}, a comprehensive multi-class image and video dataset for gastrointestinal endoscopy.
\newblock \emph{Scientific Data}, 7\penalty0 (1):\penalty0 283, 2020.

\bibitem[Alabi et~al.(2025)Alabi, Toe, Zhou, Budd, Raison, Shi, and Vercauteren]{alabi2025cholecinstanceseg}
Oluwatosin Alabi, Ko~Ko~Zayar Toe, Zijian Zhou, Charlie Budd, Nicholas Raison, Miaojing Shi, and Tom Vercauteren.
\newblock Cholecinstanceseg: A tool instance segmentation dataset for laparoscopic surgery.
\newblock \emph{Scientific Data}, 12\penalty0 (1):\penalty0 825, 2025.

\bibitem[Wagner et~al.(2023)Wagner, M{\"u}ller-Stich, Kisilenko, Tran, Heger, M{\"u}ndermann, Lubotsky, M{\"u}ller, Davitashvili, Capek, et~al.]{wagner2023comparative}
Martin Wagner, Beat-Peter M{\"u}ller-Stich, Anna Kisilenko, Duc Tran, Patrick Heger, Lars M{\"u}ndermann, David~M Lubotsky, Benjamin M{\"u}ller, Tornike Davitashvili, Manuela Capek, et~al.
\newblock Comparative validation of machine learning algorithms for surgical workflow and skill analysis with the {HeiChole} benchmark.
\newblock \emph{Medical Image Analysis}, 86:\penalty0 102770, 2023.

\bibitem[Psychogyios et~al.(2023)Psychogyios, Colleoni, Van~Amsterdam, Li, Huang, Li, Jia, Zou, Wang, Liu, et~al.]{psychogyios2023sar}
Dimitrios Psychogyios, Emanuele Colleoni, Beatrice Van~Amsterdam, Chih-Yang Li, Shu-Yu Huang, Yuchong Li, Fucang Jia, Baosheng Zou, Guotai Wang, Yang Liu, et~al.
\newblock Sar-rarp50: Segmentation of surgical instrumentation and action recognition on robot-assisted radical prostatectomy challenge.
\newblock \emph{arXiv preprint arXiv:2401.00496}, 2023.

\bibitem[Zia et~al.(2023)Zia, Bhattacharyya, Liu, Berniker, Wang, Nespolo, and Jarc]{zia2023surgical}
A.~Zia, K.~Bhattacharyya, X.~Liu, M.~Berniker, Z.~Wang, R.~Nespolo, and A.~Jarc.
\newblock Surgical tool classification and localization: results and methods from the miccai 2022 surgtoolloc challenge.
\newblock \emph{arXiv preprint arXiv:2305.07152}, 2023.

\bibitem[Allan et~al.(2020)Allan, Kondo, Bodenstedt, Leger, Kadkhodamohammadi, Luengo, Fuentes, Flouty, Mohammed, Pedersen, et~al.]{endovis18}
Max Allan, Satoshi Kondo, Sebastian Bodenstedt, Stefan Leger, Rahim Kadkhodamohammadi, Imanol Luengo, Felix Fuentes, Evangello Flouty, Ahmed Mohammed, Marius Pedersen, et~al.
\newblock 2018 robotic scene segmentation challenge.
\newblock \emph{arXiv preprint arXiv:2001.11190}, 2020.

\bibitem[Murali et~al.(2023)Murali, Alapatt, Mascagni, Vardazaryan, Garcia, Okamoto, Costamagna, Mutter, Marescaux, Dallemagne, and Padoy]{murali2023endoscapes}
Aditya Murali, Deepak Alapatt, Pietro Mascagni, Armine Vardazaryan, Alain Garcia, Nariaki Okamoto, Guido Costamagna, Didier Mutter, Jacques Marescaux, Bernard Dallemagne, and Nicolas Padoy.
\newblock The endoscapes dataset for surgical scene segmentation, object detection, and critical view of safety assessment: Official splits and benchmark.
\newblock \emph{arXiv preprint arXiv:2312.12429}, 2023.

\bibitem[Nwoye et~al.(2022)Nwoye, Yu, Gonzalez, Seeliger, Mascagni, Mutter, Marescaux, and Padoy]{nwoye2022rendezvous}
Chinedu~Innocent Nwoye, Tong Yu, Cristians Gonzalez, Barbara Seeliger, Pietro Mascagni, Didier Mutter, Jacques Marescaux, and Nicolas Padoy.
\newblock Rendezvous: Attention mechanisms for the recognition of surgical action triplets in endoscopic videos.
\newblock \emph{Medical Image Analysis}, 78:\penalty0 102433, 2022.

\bibitem[Nwoye et~al.(2025)Nwoye, Elgohary, Srinivas, Zaid, Lavanchy, and Padoy]{nwoye2023cholectrack20}
Chinedu~Innocent Nwoye, Kareem Elgohary, Anvita Srinivas, Fauzan Zaid, Jo{\"e}l~L Lavanchy, and Nicolas Padoy.
\newblock Cholectrack20: A multi-perspective tracking dataset for surgical tools.
\newblock In \emph{Proceedings of the IEEE/CVF Conference on Computer Vision and Pattern Recognition}, pages 22512--22522, 2025.

\bibitem[Ayobi et~al.(2025)Ayobi, Rodr{\'\i}guez, P{\'e}rez, Hern{\'a}ndez, Aparicio, Dessevres, Pe{\~n}a, Santander, Caicedo, Fern{\'a}ndez, et~al.]{ayobi2025pixel}
Nicol{\'a}s Ayobi, Santiago Rodr{\'\i}guez, Alejandra P{\'e}rez, Isabela Hern{\'a}ndez, Nicol{\'a}s Aparicio, Eug{\'e}nie Dessevres, Sebasti{\'a}n Pe{\~n}a, Jessica Santander, Juan~Ignacio Caicedo, Nicol{\'a}s Fern{\'a}ndez, et~al.
\newblock Pixel-wise recognition for holistic surgical scene understanding.
\newblock \emph{Medical Image Analysis}, page 103726, 2025.

\bibitem[Rueckert et~al.(2024)]{rueckert2024phakir}
Tobias Rueckert et~al.
\newblock {PhaKIR} challenge: Phase, keypoint and instrument recognition in laparoscopic cholecystectomy videos.
\newblock MICCAI Endoscopic Vision Challenge, 2024.

\bibitem[Allan et~al.(2021)Allan, Mcleod, Wang, Rosenthal, Hu, Gard, Eisert, Fu, Zeffiro, Xia, et~al.]{allan2021stereo}
Max Allan, Jonathan Mcleod, Congcong Wang, Jean-Claude Rosenthal, Zhenglei Hu, Niklas Gard, Peter Eisert, Ke~Xue Fu, Trevor Zeffiro, Wenyao Xia, et~al.
\newblock Stereo correspondence and reconstruction of endoscopic data challenge.
\newblock \emph{arXiv preprint arXiv:2101.01133}, 2021.

\bibitem[de~Jong et~al.(2026)de~Jong, Li, Jaspers, van Jaarsveld, Kuiper, Badaloni, van Hillegersberg, Ruurda, van~der Sommen, Pluim, et~al.]{de2026towards}
Ronald~LPD de~Jong, Yiping Li, Tim~JM Jaspers, Romy~C van Jaarsveld, Gino~M Kuiper, Franco Badaloni, Richard van Hillegersberg, Jelle~P Ruurda, Fons van~der Sommen, Josien~PW Pluim, et~al.
\newblock Towards effective surgical representation learning with dino models.
\newblock In \emph{Medical Imaging with Deep Learning}, 2026.

\bibitem[Wu et~al.(2026)Wu, Holm, Chen, Wang, Hu, Ye, Zang, Xu, Zhou, Wu, et~al.]{wu2026surgmotion}
Jinlin Wu, Felix Holm, Chuxi Chen, An~Wang, Yaxin Hu, Xiaofan Ye, Zelin Zang, Miao Xu, Lihua Zhou, Huarui Wu, et~al.
\newblock Surgmotion: A video-native foundation model for universal understanding of surgical videos.
\newblock \emph{arXiv preprint arXiv:2602.05638}, 2026.

\bibitem[Perez et~al.(2026)Perez, Nwoye, Kermani, Mohareri, and Jamal]{perez2026surglavi}
Alejandra Perez, Chinedu Nwoye, R.~R. Kermani, Omid Mohareri, and Muhammad~Abdullah Jamal.
\newblock Surglavi: Large-scale hierarchical dataset for surgical vision--language representation learning.
\newblock \emph{Medical Image Analysis}, page 103982, 2026.

\bibitem[Schoeffmann et~al.(2018)Schoeffmann, Husslein, Kletz, Petscharnig, M{\"u}nzer, and Beecks]{schoeffmann2018video}
Klaus Schoeffmann, Heinrich Husslein, Sabrina Kletz, Stefan Petscharnig, Bernd M{\"u}nzer, and Christian Beecks.
\newblock Video retrieval in laparoscopic video recordings with dynamic content descriptors.
\newblock \emph{Multimedia Tools and Applications}, 77\penalty0 (13):\penalty0 16813--16832, 2018.
\newblock \doi{10.1007/s11042-017-5252-2}.

\bibitem[Stilz et~al.(2026)Stilz, Arboit, Srivastav, {CAMMA International Surgical Partners}, Marescaux, Alfieri, Mascagni, Navab, and Padoy]{stilz2026surge}
Florian~Philipp Stilz, Lorenzo Arboit, Vinkle Srivastav, {CAMMA International Surgical Partners}, Jacques Marescaux, Sergio Alfieri, Pietro Mascagni, Nassir Navab, and Nicolas Padoy.
\newblock A surgical foundation model reveals task-dependent label efficiency.
\newblock \emph{arXiv preprint arXiv:2609.31821}, 2026.

\end{thebibliography}

\appendix
\section{Extended Data}
\section{EXTENDED DATA}

\setcounter{figure}{0}
\setcounter{table}{0}
\captionsetup[figure]{name={Extended Data Fig.}}
\captionsetup[table]{name={Extended Data Table}}

\begin{figure}[htbp]
  \centering
  \includegraphics[width=\linewidth]{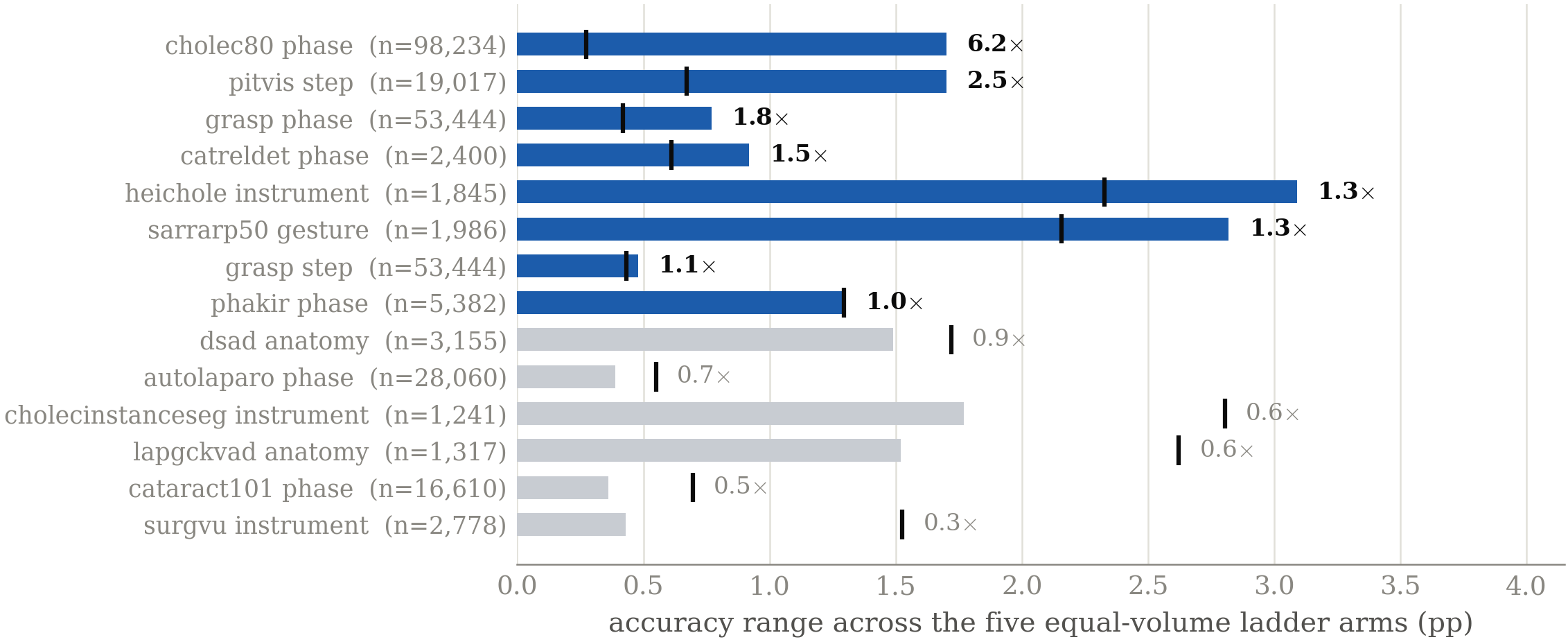}
  \caption{\textbf{Sensitivity to pretraining data composition.} Accuracy ranges across five equal-volume pretraining configurations for fourteen downstream tasks. Vertical markers indicate twice the frame-level binomial standard error, with annotations showing the ratio between the observed range and this threshold. Eight tasks exceed the threshold (blue), while six show insufficient variation to distinguish between configurations (grey). }
  \label{edfig:resolution}
\end{figure}

\begin{figure}[htbp]
  \centering
  \includegraphics[width=\linewidth]{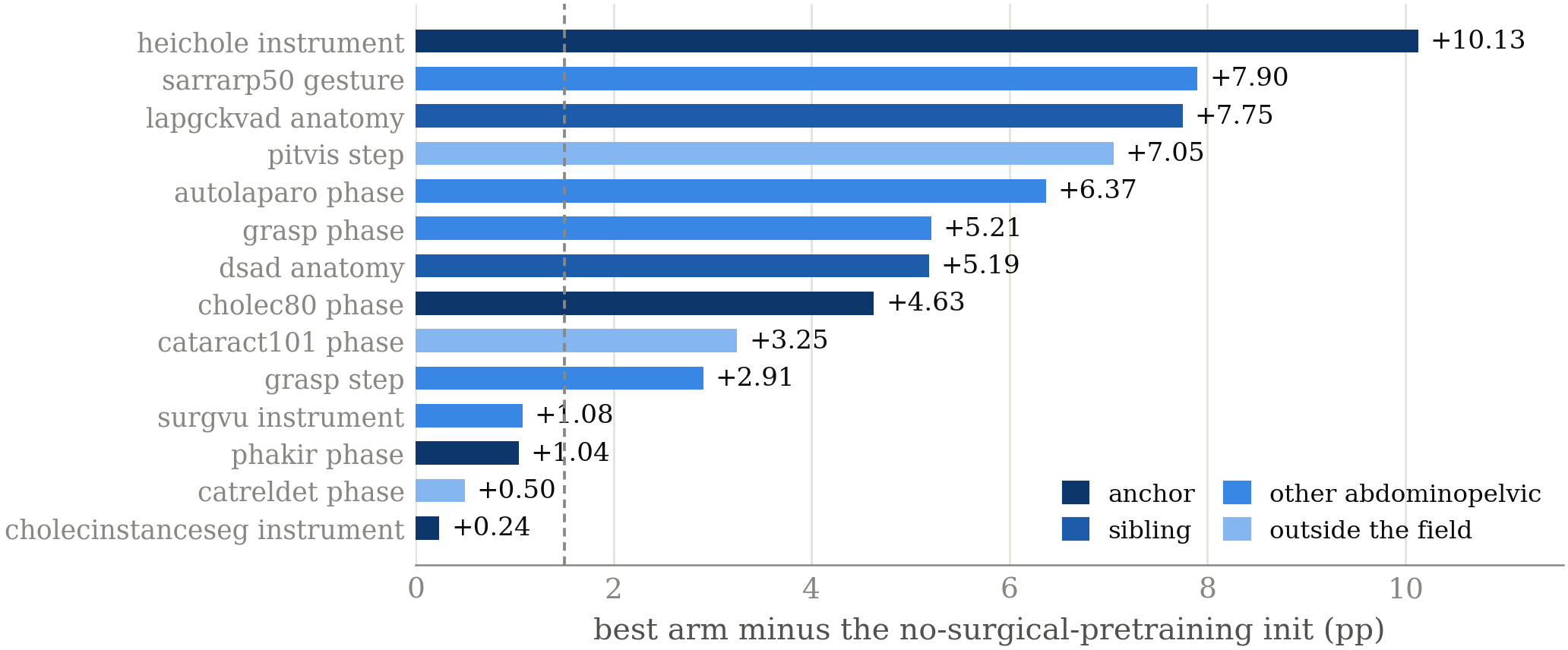}
  \caption{\textbf{Performance gains from surgical continual pretraining.} Improvement of the best-performing pretraining configuration over the original DINOv3 initialization across fourteen downstream tasks, grouped by surgical domain. Ten tasks achieve gains greater than 1.5 percentage points (dashed line), while four show limited improvements. The variation across tasks highlights the different benefits of surgical pretraining depending on the downstream application.}
  \label{edfig:gain}
\end{figure}

\begin{table*}[htbp]\centering\footnotesize\setlength{\tabcolsep}{2pt}
\caption{\textbf{Per-task linear-probe scores across the released \methodname family}
(ViT-S/16, ViT-B/16, ViT-L/16; the released flagship is the ViT-L configuration reported in
the main text). Metrics as in the main-text benchmark table; -- = not evaluated.}
\label{edtab:family}
\begin{tabular}{lcccccccccccccccccccc}
\toprule
Model & ChPh & AuPh & MbPh & MbSt & PitV & HypK & ChIS & HeiC & SARr & SgVU & LapG & AuSg & EnSg & CVS & Trip & ChoTr & GrSP-P & GrSP-S & Phak & Vid \\
\midrule
\methodname ViT-S & 78.1 & 69.2 & 82.1 & 70.6 & 71.6 & 87.5 & 59.8 & 45.9 & 39.0 & 24.7 & 67.6 & 74.5 & 33.1 & 38.3 & 15.1 & 75.2 & 62.3 & 49.5 & 66.7 & -- \\
\methodname ViT-B & 80.7 & 72.9 & 84.9 & 74.0 & 74.5 & 88.9 & 67.0 & 50.7 & 40.8 & 23.9 & 72.1 & 76.0 & 37.1 & 43.7 & 17.1 & 78.4 & 65.9 & 53.5 & 70.1 & 50.0 \\
\methodname ViT-L & 83.5 & 76.3 & 86.4 & 75.7 & 74.5 & 90.0 & 72.7 & 56.7 & 43.0 & 26.9 & 73.7 & 79.7 & 39.3 & 50.4 & 20.5 & 84.7 & 67.3 & 53.2 & 75.2 & 53.1 \\
\bottomrule
\end{tabular}
\end{table*}

\begin{table}[htbp]
\centering
\caption{Pre-training dataset statistics and composition.}
\label{edtab:dataset}
\setlength{\tabcolsep}{6pt}
\begin{tabular}{lrrl}
\toprule
\textbf{Category} & \textbf{Videos} & \textbf{Hours} & \textbf{Notes} \\
\midrule
\multicolumn{4}{l}{\textit{Surgical type breakdown}} \\
\midrule
Hernia repair              & 1,098 & $\sim$237 h & \\
Nephrectomy / Urology      & 1,039 & $\sim$224 h & \\
Colectomy / Colorectal     & 1,036 & $\sim$224 h & \\
Cholecystectomy            &   439 & $\sim$95 h  & \\
Thoracic                   &   421 & $\sim$91 h  & \\
Proctology / Rectal        &   414 & $\sim$89 h  & \\
Gynecology                 &   413 & $\sim$89 h  & \\
Gastrectomy / Bariatric    &   384 & $\sim$83 h  & \\
Liver / Biliary            &   364 & $\sim$79 h  & \\
Fundoplication / GERD      &   303 & $\sim$65 h  & \\
Appendectomy               &   294 & $\sim$63 h  & \\
Pancreas                   &   198 & $\sim$43 h  & \\
Splenectomy                &   198 & $\sim$43 h  & \\
Lymph node dissection      &   112 & $\sim$24 h  & \\
Vascular                   &    47 & $\sim$10 h  & \\
Robotic training           &    25 & $\sim$5 h   & \\
Uncategorized              & 5,180 & $\sim$1450 h & \\
\bottomrule
\end{tabular}
\end{table}

\begin{figure}[htbp]
    \centering
    \includegraphics[width=\linewidth]{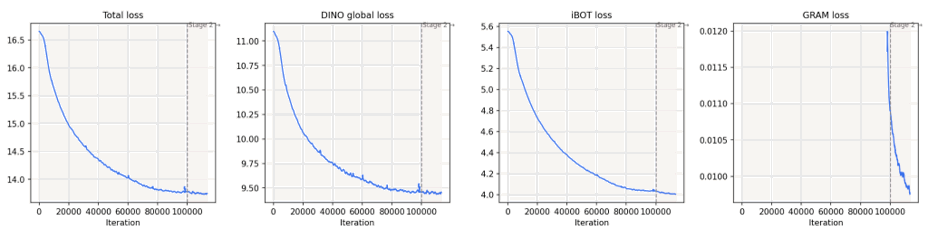}
    \caption{Plots illustrate the joint optimization of the DINO (global distillation), iBOT (masked patch reconstruction), and GRAM objectives. The rapid initial descent followed by stable convergence indicates the model's successful adaptation from general visual priors to the complex spatial and colorimetric distributions of the surgical domain, achieving high-quality representation alignment without experiencing dense feature collapse.}
    \label{edfig:loss}
\end{figure}

\end{document}